\documentclass[letterpaper, 10 pt, conference]{ieeeconf}  

\IEEEoverridecommandlockouts                              

\usepackage{graphicx}
\usepackage{amsmath,amssymb,amsfonts}
\usepackage{comment}
\usepackage{caption}
\usepackage{subcaption}
\usepackage{multirow}
\usepackage{float}    
\usepackage{placeins} 
\usepackage{booktabs}
\usepackage{svg}
\usepackage{cite}
\usepackage{hyperref}
\usepackage{dsfont}
\usepackage{tikz}
\usetikzlibrary{arrows.meta,positioning,fit,calc,shapes.geometric, backgrounds}
\usepackage[table]{xcolor}
\usepackage[ruled,vlined,linesnumbered,noend]{algorithm2e}
\usepackage{pifont}
\usepackage{array} 
\newcolumntype{P}[1]{>{\centering\arraybackslash}p{#1}}

\usepackage{bm}
\definecolor{myred}{RGB}{160,40,35}
\definecolor{mygreen}{RGB}{50,140,50} 
\usepackage[acronym]{glossaries}
\newacronym{ad}{AD}{Autonomous Driving}
\newacronym{e2e}{E2E}{End-to-End}
\newacronym{il}{IL}{imitation learning}
\newacronym{rl}{RL}{Reinforcement Learning}
\newacronym{morl}{MORL}{Multi-Objective Reinforcement Learning}
\newacronym{hrl}{HRL}{Hierarchical Reinforcement Learning}
\newacronym{iqn}{IQN}{Implicit Quantile Network}
\newacronym{llm}{LLM}{Large Language Model}
\newacronym{dqn}{DQN}{Deep Q-Network}
\newacronym{momdp}{MOMDP}{Multi-objective markov decision process}
\newacronym{pmomdp}{Pr-MOMDP}{Preordered MOMDP}
\newacronym{kl}{KL}{Kullback–Leibler}
\newacronym{cvar}{CVaR}{conditional value-at-risk}
\newacronym{mv}{MV}{mean variance}
\newacronym{qd}{QD}{quantile dominance}
\newacronym{dqd}{DQD}{directional quantile distance}
\newacronym{td}{TD}{temporal difference}
\newacronym{maiqn}{MA-IQN}{mean aggregated IQN}

\usepackage{eso-pic}
\newcommand{\mycopyrighttext}{%
  \footnotesize
  \noindent
  \textcopyright~2026 IEEE. Personal use of this material is permitted. Permission from IEEE must be obtained for all other uses, in any current or future media, including reprinting/republishing this material for advertising or promotional purposes, creating new collective works, for resale or redistribution to servers or lists, or reuse of any copyrighted component of this work in other works.\\
 IEEE International Conference on Robotics and Automation (ICRA) - June 1-5, 2026.
}

\AtBeginDocument{%
  \AddToShipoutPicture*{%
    \AtPageUpperLeft{%
      \put(55, -40){
        \parbox{\dimexpr\textwidth-4pt\relax}{\raggedright \mycopyrighttext}
      }%
    }
  }
}

\title{\LARGE \bf Beyond Scalar Rewards: Distributional Reinforcement Learning with Preordered Objectives for Safe and Reliable Autonomous Driving}
\author{Ahmed Abouelazm\textsuperscript{\textasteriskcentered}$^{1,2}$, Jonas Michel\textsuperscript{\textasteriskcentered}$^{2}$, Daniel Bogdoll$^{1,2}$, Philip Schörner$^{1,2}$, and J. Marius Zöllner$^{1,2}$
\thanks{\textasteriskcentered~These authors contributed equally to this work}%
\thanks{$^{1}$Authors are with the FZI Research Center for Information Technology, Germany {\tt\small name@fzi.de}}%
\thanks{$^{2}$Authors are with the Karlsruhe Institute of Technology, Germany}
}

\begin{document}
\bstctlcite{IEEEexample:BSTcontrol}
\maketitle
\thispagestyle{empty}
\pagestyle{empty}

\begin{abstract}
    Autonomous driving involves multiple, often conflicting objectives such as safety, efficiency, and comfort. In reinforcement learning (RL), these objectives are typically combined through weighted summation, which collapses their relative priorities and often yields policies that violate safety-critical constraints. To overcome this limitation, we introduce the Preordered Multi-Objective MDP (Pr-MOMDP), which augments standard MOMDPs with a preorder over reward components. This structure enables reasoning about actions with respect to a hierarchy of objectives rather than a scalar signal. To make this structure actionable, we extend distributional RL with a novel pairwise comparison metric, Quantile Dominance (QD), that evaluates action return distributions without reducing them into a single statistic. Building on QD, we propose an algorithm for extracting optimal subsets, the subset of actions that remain non-dominated under each objective, which allows precedence information to shape both decision-making and training targets. Our framework is instantiated with Implicit Quantile Networks (IQN), establishing a concrete implementation while preserving compatibility with a broad class of distributional RL methods. Experiments in Carla show improved success rates, fewer collisions and off-road events, and deliver statistically more robust policies than IQN and ensemble-IQN baselines. By ensuring policies respect rewards preorder, our work advances safer, more reliable autonomous driving systems.

\end{abstract}
\section{Introduction}
\label{sec:Introduction}
\gls{e2e} learning has emerged as a compelling paradigm for \gls{ad}, directly mapping raw sensory input to vehicle actions within a unified model~\cite{tampuu2020survey}. By bypassing handcrafted intermediate stages, \gls{e2e} approaches mitigate error accumulation and enable scalable, data-driven driving policies~\cite{endtoendreview}. Unlike modular pipelines, which rely on carefully engineered perception and decision-making components, \gls{e2e} systems reduce error propagation between modules and allow joint optimization of perception and control. This streamlines learning and lowers reliance on manual design~\cite{hu2023planning}.

While \gls{il} is effective for acquiring basic driving skills, \gls{rl} offers distinct advantages by optimizing behavior through direct interaction with the environment~\cite{lu2023imitation}. By maximizing cumulative rewards, \gls{rl} agents can adapt to the dynamic and uncertain conditions of real-world traffic~\cite{kiran2021deep}. At the core of this process lies the reward function, which specifies driving objectives such as safety, efficiency, and comfort, and thus directly governs the quality of the learned policy~\cite{chen2024end}.

\textbf{Research Gap.} Designing reward functions for \gls{ad} is inherently complex, as they must capture multiple, often conflicting objectives, such as safety, efficiency, and comfort, while also reflecting their relative priorities~\cite{abouelazm2024review}. A common practice is to collapse these objectives into a single scalar reward, typically through naive or weighted summation~\cite{kiran2021deep}. However, such formulations are difficult to tune, highly context-dependent, and have been shown to produce undesirable behaviors when trade-offs are misaligned, such as prioritizing comfort or efficiency at the expense of safety~\cite{knox2023reward}.

To address these deficiencies, research in \gls{morl} has proposed representing a reward as a vector of distinct objectives~\cite{deshpande2021navigation}. This allows the agent to learn separate value estimates per objective. However, decision-making still relies on weighted aggregation of these estimates~\cite{juozapaitis2019explainable}, which limits the learned policy's ability to preserve the intended priority of objectives.

Alternative approaches introduce hierarchical rewards inspired by rulebooks~\cite{censi2019liability}, where relations among objectives are explicitly encoded~\cite{bogdoll2024informed, Balancing_Abouelazm_2025}. While these approaches provide interpretability and structured priorities at the reward-design level, agents are still trained with scalar rewards, restricting agents' ability to disentangle the contribution of each objective.

Therefore, a gap remains in developing \gls{rl} agents that can semantically represent multiple objectives and enforce their relative priorities within the learning process itself. Such a formulation enables safety-critical objectives to guide both learning and decision-making more reliably, leading to safer driving behavior.

\textbf{Contribution.} This paper presents a framework that incorporates preorder relations, capturing the relative priority between objectives, directly into \gls{rl} agents. In this way, objective priorities are respected during both training and inference. The key contributions of this work are:
\begin{itemize}
    \item \textbf{Preordered Multi-Objective MDP (Pr-MOMDP):} We extend MOMDPs with preorder relations, providing a formulation that enables reasoning about actions with respect to prioritized objectives.  

    \item \textbf{Quantile-based action comparison:} Building on the Pr-MOMDP formulation, we propose Quantile Dominance (QD), a distributional metric that compares full return distributions to derive pairwise action relations.  

    \item \textbf{Optimal subsets for decision-making:} Leveraging QD, we extract optimal action subsets, non-dominated actions under each objective, and integrate them into both action selection and training updates, ensuring higher-priority objectives consistently shape policy learning.  
\end{itemize}

\section{related work}
The integration of multi-objective and hierarchical reward structures into \gls{rl} policies is a critical area of research. While significant progress has been made, major challenges remain. Experiments in the complex domain of \gls{ad} have revealed that current approaches insufficiently respect reward structures, leading to undesirable behavior~\cite{knox2023reward}. In this section, we first introduce classical reward structures and their challenges in the context of \gls{ad}, and subsequently introduce prior work in \gls{morl} and \gls{hrl} that aims to handle such complex reward structures.
\subsection{Reward Design in Autonomous Driving}
\gls{ad} is a complex domain with a multitude of often conflicting objectives~\cite{abouelazm2024review}. This makes it challenging to manually design reward functions and can often lead to insufficient performance~\cite{skalseDefiningCharacterizingReward2022}. Knox et al.~\cite{knox2023reward} identified several challenges in the manual design of reward functions, such as undesired risk tolerance or preference orderings that do not align with human judgment. These findings are confirmed by a large-scale survey by Abouelazm et al.~\cite{abouelazm2024review}, highlighting that most reward functions in the literature utilize different individual reward terms but aggregate them into a scalar output, eliminating context awareness and relative ordering.

To avoid error-prone manual reward designs, another line of work proposes the automated generation of a reward function based on \glspl{llm}~\cite{hanAutoRewardClosedLoopReward2024}. However, this approach still collapses multiple reward terms into a single scalar, not solving the core issue of classical reward structures. Finally, some works address the runtime adoption of driving behaviors by introducing priors on driving aspects, such as comfort or aggressiveness~\cite{joseph_dreamtodrive_2024, surmann2025multiobjectivereinforcementlearningadaptable}. Here, classically engineered reward terms are combined with prior conditions so agents can display different behaviors during inference without re-training. These approaches emphasize certain aspects of total reward rather than integrating hierarchies to address complex reward structures. 

\subsection{Multi-Objective RL and Hierarchical Rewards.}
As classical reward structures struggle to address scenarios that require hierarchical or multi-objective reward signals, this section presents advancements in \gls{morl} and \gls{hrl}.

Rather than just decomposing components of the reward function, several works adapt model architectures by introducing multi-branch networks for individual reward components~\cite{Seijen_hybrid_reward_17, yuan_multireward_2019, liUrbanDrivingMultiObjective2019,juozapaitis2019explainable, jinHybridActionBased2025,macglashanValueFunctionDecomposition}. While these concepts have demonstrated performance improvements and can be used to dynamically adjust weights during runtime~\cite{jamilDynamicWeightbasedMultiObjective2022, alegreAMORAdaptiveCharacter2025}, an artificial bottleneck is introduced by merging the resulting Q-values into a singular value for training. 

Differently, Deshpande et al.~\cite{deshpande2021navigation} use a \gls{dqn} per reward objective and generate a list of acceptable actions for each. However, the action selection is based on sequential filtering and ordering, which is error-prone and cannot capture the full range of relations between objectives. Several works have combined distributional RL with multi-dimensional rewards~\cite{linDistributionalRewardDecomposition2019,zhangDistributionalReinforcementLearning2021, caiDistributionalParetoOptimalMultiObjective, wiltzerFoundationsMultivariateDistributional}. However, they utilize simple, unstructured rewards, which are not suitable to address complex real-world applications such as \gls{ad}.

Only a few works integrate complex reward structures in \gls{rl} applications. Bogdoll et al.~\cite{bogdoll2024informed} proposed a rulebook-based~\cite{censi2019liability} and situation-aware reward function that showed performance improvements in traffic scenarios that required controlled rule exceptions. Abouelazm et al.~\cite{Balancing_Abouelazm_2025} similarly designed rulebook-based rewards with a novel risk term and a normalization scheme that assigns weights according to the hierarchy level of each reward term. However, both approaches collapse the structured reward into a single scalar, limiting their ability to fully exploit the hierarchy.

Unlike prior approaches that collapse objectives into a single scalar (Fig.~\ref{fig:sub1}), aggregate value estimates without preserving preorder (Fig.~\ref{fig:sub2}), or rely on unstructured rewards, our approach preserves preorder between objectives, leverages distributional value estimates for robust action comparison, and encodes reward hierarchies directly into the learning process.
\section{Methodology}
In this section, we formalize our approach, illustrated in Fig.~\ref{fig:sub3}, to incorporating reward preorder into \gls{rl}. We first extend the MOMDP with a preorder relation, referred to as precedence among objectives, to capture hierarchical structure (Section~\ref{sec:problem_formulation}). 
We then introduce a distributional metric for action comparison and a preorder-guided action selection framework that adapts both architecture and training to respect priorities (Section~\ref{sec:action_relations}).

\subsection{Problem Formulation}
\label{sec:problem_formulation}
\gls{morl} is typically formalized through \gls{momdp}, $ \mathcal{M}_{\text{MOMDP}} = \left\langle S, A, P, \mathcal{R}, \gamma \right\rangle$, where $S$ is a finite set of states, $A$ a finite set of actions, and $P(s' \mid s,a)$ denotes the transition probability from state $s$ to state $s'$ under action $a$.

For $N$ objectives, the reward function is a vector given by $\mathcal{R}: S \times A \rightarrow \mathbb{R}^N$. For any $(s,a) \in S \times A$, the vectorized reward is realized as $\mathcal{R}(s,a) = \{ r_i(s,a) \}_{i=1}^N$, where $r_i(s,a)$ denotes the reward for objective $i$. The discount factors are similarly defined as $\gamma = \{ \gamma_i\}_{i=1}^N$. 

Existing \gls{momdp} formulations typically handle multiple objectives either by treating all reward components as equally weighted and aggregating them into a single signal~\cite{yuan2019multi} or by enforcing a strict lexicographic order~\cite{deshpande2021navigation}. Both approaches impose rigid constraints that limit the framework’s ability to capture more flexible relations among objectives.

\begin{figure}[t]
    \centering
\begin{subfigure}[t]{0.43\columnwidth}
    \centering
    \begin{tikzpicture}[
        >=Stealth,
        rule/.style={draw, rounded corners=6pt, minimum width=11mm, minimum height=5.5mm, align=center}
    ]
        \node[rule] (r1) at (0,1.8) {$r_1$};
        \node[rule] (r2) at (0,0.6) {$r_2$};
        \node[rule] (r3) at (0,-0.6) {$r_3$};
        \node[rule] (r4) at (0,-1.8) {$r_4$};

        \draw[->, shorten >=2pt, shorten <=2pt] (r4.north) -- (r3.south);
        \draw[->, shorten >=2pt, shorten <=2pt] (r3.north) -- (r2.south);
        \draw[->, shorten >=2pt, shorten <=2pt] (r2.north) -- (r1.south);
    \end{tikzpicture}
    \caption{Lexicographic Reward}
    \end{subfigure}
    \hspace{0.02\columnwidth}
    \begin{subfigure}[t]{0.43\columnwidth}
        \centering
        \begin{tikzpicture}[
            >=Stealth,
            node distance=1.2cm,
            rule/.style={draw, rounded corners=6pt, minimum width=11mm, minimum height=5.5mm, align=center}
        ]
            \node[rule] (r1) at (0,2.4) {$r_1$};
            \node[rule] (r2) at (-1.5,0.8) {$r_2$};
            \node[rule] (r3) at (1.5,0.8) {$r_3$};
            \node[rule] (r4) at (0, -0.8) {$r_4$};
            \draw[->, shorten >=2pt, shorten <=2pt] (r4.north) -- (r2.south);
            \draw[->, shorten >=2pt, shorten <=2pt] (r4.north) -- (r3.south);
            \draw[->, shorten >=2pt, shorten <=2pt] (r2.north) -- (r1.south);
            \draw[->, shorten >=2pt, shorten <=2pt] (r3.north) -- (r1.south);
        \end{tikzpicture}
        \caption{Partially Ordered Reward}
    \end{subfigure}
    \caption{Examples of lexicographic and partial order rewards}
    \label{fig:example_rulebooks}
\end{figure}
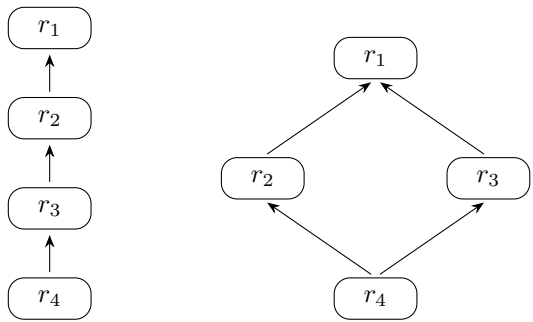

To address this limitation, we introduce the \gls{pmomdp}, an extension of \gls{momdp} that incorporates a pre-order relation $\succeq$ over reward components. This extension preserves the vectorized reward structure while enabling comparisons that respect the hierarchy among objectives. In contrast to rulebooks~\cite{censi2019liability}, which use $\preceq$ because they operate on costs to be minimized, we employ $\succeq$ since our formulation is reward-based and maximizes returns. The proposed formulation of \gls{pmomdp} is given in Eq.~\ref{eq:pmomdp}.
\begin{equation}
\mathcal{M}_{\text{Pr-MOMDP}} =  \mathcal{M}_{\text{MOMDP}} + \langle \,\succeq \, \rangle = \langle S, A, P, \mathcal{R}, \gamma, \succeq \rangle
\label{eq:pmomdp}
\end{equation}
For any $r_i, r_j \in \mathcal{R}$, the relation $r_i \succeq r_j$ indicates that the reward component $r_i$ has a higher priority than $r_j$. The introduction of a pre-order allows reward relations to be represented flexibly as directed graphs. Figure~\ref{fig:example_rulebooks} illustrates two instances: a total order (lexicographic) with $r_1 \succeq r_2 \succeq r_3 \succeq r_4$, and a partial order in which $r_2$ and $r_3$ remain incomparable. Such flexibility is essential for capturing both strict hierarchies and more general priority structures that arise in multi-objective decision-making.

To address the complexities of \gls{ad}, we extend the formulation in Eq.~\ref{eq:pmomdp} from the fully observable case to the more challenging partially observable setting. Here, the agent interacts with the environment through observations $o \in O$ generated by a sensor model. This extension leaves the preorder relation $\succeq$ unaffected, as prioritization among reward components is independent of the observation process.

\subsection{Action Relations from Rewards Preorder}
The introduction of a precedence relation not only structures the reward components themselves but also induces a relational semantics among actions. Building on the relations introduced in~\cite{halder2025sampling}, we adapt them to the proposed Pr-MOMDP setting: given two actions $a, a' \in A$ with corresponding reward components $\mathcal{R}(s,a), \mathcal{R}(s,a') \in \mathbb{R}^N$ and a preorder relation $\succeq$ over objectives, we define:
\begin{itemize}
    \item \textbf{Dominance:}
    $a$ dominates $a'$ if there exists an objective $r_j$ satisfying 
    $r_j(s,a) > r_j(s,a')$, and for any objective with 
    $r_i(s,a') > r_i(s,a)$, it holds that $r_j \succ r_i$ under $\succeq$.

    \item \textbf{Indifference:}  
    $a$ and $a'$ are indifferent if neither $a$ dominates $a'$ 
    nor $a'$ dominates $a$.

    \item \textbf{Incomparability:}  
    $a$ and $a'$ are incomparable if there exist objectives 
    $r_i$ and $r_j$ such that $r_i(s,a) > r_i(s,a')$ and $r_j(s,a') > r_j(s,a)$, and neither objective is comparable (ordered above the other) under $\succeq$.
\end{itemize}

Precedence-based relations provide a meaningful way to compare actions in terms of their reward vectors, but \gls{rl} agents do not act directly on rewards. Instead, decisions are guided by value functions that estimate expected return over time. To enable agents to benefit from the semantic structure of rewards, we address how precedence-based action relations can be extended into the value-function space. The next section develops a comparison algorithm that integrates these relations into learning, allowing objective hierarchies to guide action evaluation and policy learning.
\subsection{Preorder-guided Action Selection}
\label{sec:action_relations}
\subsubsection{Agent Architecture}
Representing multiple value functions within the agent architecture raises design challenges. Factored-state approaches~\cite{li2018urban} assign each objective to a separate subset of the state, but this requires handcrafted features and is infeasible when learning directly from raw sensor data. Using the full state with separate networks per objective~\cite{deshpande2021navigation} avoids hand-design but duplicates computation and prevents objectives from benefiting from shared representations~\cite{ruder2017overview}.

\begin{figure*}[t]
    \centering
    \begin{subfigure}[b]{0.3\textwidth}
        \centering
        \includegraphics[height=7cm]{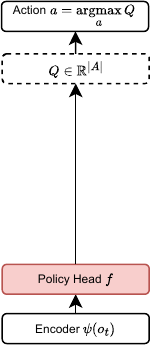}
        \caption{Single-Head architectures~\cite{Balancing_Abouelazm_2025,bogdoll2024informed}, where all objectives are entangled in a single policy head $f$ without the ability to separate them. }
        \label{fig:sub1}
    \end{subfigure}
    \hfill
    \begin{subfigure}[b]{0.3\textwidth}
        \centering
        \includegraphics[height=7cm]{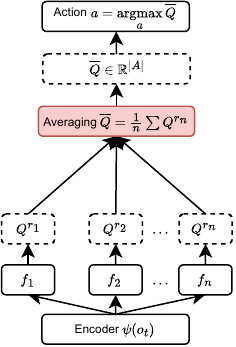}
        \caption{Multi-Head  architectures~\cite{juozapaitis2019explainable,Seijen_hybrid_reward_17}, which learns one head per objective $r_i$ but collapses decision-making to the mean value estimate.}
        \label{fig:sub2}
    \end{subfigure}
    \hfill
    \begin{subfigure}[b]{0.3\textwidth}
        \centering
        \includegraphics[height=7cm]{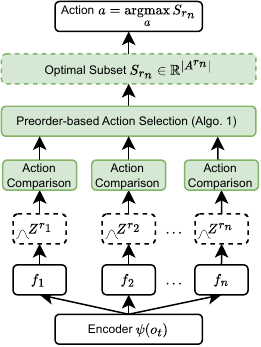}
        \caption{Our Pr-IQN with a novel action comparison and selection algorithm to utilize all available information and preserve a given preorder.}
        \label{fig:sub3}
    \end{subfigure}
    \caption{Comparison of two classical architectures (a, b) with our Pr-IQN approach (c), shown during inference given observations $o_t$. Information bottlenecks are highlighted in \textcolor{myred}{red} and novel components for full information utilization in \textcolor{mygreen}{green}. Compared to classical approaches, Pr-IQN leverages distributions $Z^{r_n}$ to select actions that respect a given preorder.}
    \label{fig:architectures}
    \vspace{-0.4cm}
\end{figure*}

To address these limitations, we adopt a multi-head architecture: observations are encoded into a common latent representation that captures task-relevant features, which is then passed to multiple heads, as demonstrated in Fig.~\ref{fig:sub3}. Each head outputs a value estimate for its corresponding objective $r_i \in \mathcal{R}$, allowing new objectives to be added simply by introducing an additional head. This design improves scalability, 
and leverages shared features while maintaining objective-specific value predictions.




\subsubsection{Value Functions Estimation}
Strictly applying precedence at the level of value estimates raises important challenges. Previous rulebook approaches~\cite{helou2021reasonable} rely on discrete, boolean comparisons that assume rewards can be evaluated as satisfied or violated. Value functions, by contrast, often have large magnitudes, are noisy, and fluctuate during training. Enforcing strict prioritization in this setting can lead to undesirable outcomes. For example, strict prioritization may lead the agent to favor a negligible gain in clearance over significant progress, resulting in overly conservative behavior such as remaining stationary. Such brittleness highlights the need for a more tolerant evaluation mechanism that accounts for the uncertainty in value estimates and can capture precedence in a distributional form.

To address these limitations, we adopt a distributional~\gls{rl} approach. Specifically, we use quantile-based value function estimates inspired by \gls{iqn}~\cite{dabney2018implicit}, combined with a distribution-aware metric for pairwise action comparison (Section~\ref{sec:pairwise_compare}) to enable more tolerant evaluation of actions under the same objective $r_i$. These comparisons then inform a preorder-based action selection algorithm (Algo.~\ref{alg:precedence_filter}) that maintains, for each objective, the subset of non-dominated actions consistent with the hierarchy, denoted the optimal subset.

\gls{iqn} models the entire quantile function, treating the quantiles $\tau \in [0,1]$ as a continuous random variable. This allows the network to approximate the inverse cumulative distribution function (inverse CDF $F^{-1}$) of the return distribution, as expressed in Eq.~\ref{eq:inverse_cdf}. For each objective $r_i$, the network outputs a matrix $Z^{r_i}(o_t,a) \in \mathbb{R}^{ \left | \tau \right | \times \left |A \right |}$, where each row corresponds to a sampled quantile $\tau$ and each column to an action $a$.
\begin{equation}
    Z^{r_i}_{\tau}(o_t,a)
    \;\approx\;F^{-1}_{\,Z^{r_i}(o_t,a)}(\tau),
    \quad \tau \sim \mathcal{U}[0,1]
    \label{eq:inverse_cdf}
\end{equation}
\subsubsection{Distribution-aware Pairwise Comparison}
\label{sec:pairwise_compare}
This section focuses on comparing actions using the full return distribution of a reward component $r_i$. Previous works collapse the distribution into a single statistic, such as \gls{cvar}~\cite{dabney2018implicit} or \gls{mv}~\cite{theate2023risk}, thereby discarding distributional structure and increasing sensitivity to noise. In contrast, we propose \gls{qd}, a distribution-aware metric that compares action distributions to a quantile-wise ideal reference.

To ensure such comparisons are robust to estimation noise and consistent across objectives, we enforce scale invariance by normalizing quantile estimates per objective using Z-score normalization, denoted by $\tilde{Z}^{r_i}$. We then define the ideal distribution as the maximum return across all actions at each quantile $\tau_k$, as shown in Eq.~\ref{eq:ideal_dis}.
\begin{equation}
Z^{*,r_i}_{\tau_k} = \max_{a \in A} \tilde{Z}^{\,r_i}_{\tau_k}(a)
\label{eq:ideal_dis}
\end{equation}
Accordingly, the quality of an action is measured by its Wasserstein-1 distance to this ideal profile, as given in Eq.\ref{eq:wasser_distance}. Since smaller distances indicate stronger alignment with the quantile-wise optimum, we define the scalar action score $\text{score}^{r_i}(a) = -\widehat{W}^{\,r_i}_1(a)$, such that higher values correspond to stronger quantile dominance.
\begin{equation}
\widehat{W}^{\,r_i}_1(a)
=
\frac{1}{|\tau|}
\sum_{\tau_k \in \tau}
\left|
\tilde{Z}^{r_i}_{\tau_k}(a)
-
Z^{*,r_i}_{\tau_k}
\right|.
\label{eq:wasser_distance}
\end{equation}
Finally, we define \gls{qd} between two actions as demonstrated in Eq.~\ref{eq:quantile_dominance}, which quantifies the directional difference in quantile dominance. By construction, \gls{qd} is asymmetric, i.e., $\mathrm{QD}_{\,a \to a'} \neq \mathrm{QD}_{\,a' \to a}$.
\begin{equation}
\mathrm{QD}^{r_i}_{a \to a'} = \text{score}^{r_i}(a) - \text{score}^{r_i}(a').
\label{eq:quantile_dominance}
\end{equation}
To avoid overly strict action comparisons, we introduce a tolerance parameter $\epsilon_{r_i} \in \mathbb{R}$ for each objective. Two actions $a$ and $a'$ are deemed \emph{indifferent} if $|\mathrm{QD}^{r_i}_{a \to a'}| \leq \epsilon_{r_i}$. Action $a$ \emph{dominates} $a'$ if $\mathrm{QD}^{r_i}_{a \to a'} > \epsilon_{r_i}$, and is \emph{dominated by} $a'$ otherwise. The \gls{qd} procedure provides a principled way to assign pairwise relations between actions under a single objective $r_i$. In the next section, we extend it from individual objectives to the full pre-order structure over rewards.

\subsubsection{Preorder Traversal and Action Selection}
\label{sec:global_compare}
In contrast to rulebook planners~\cite{halder2025sampling} that require exhaustive evaluation over all objectives and actions, our algorithm is more efficient. It operates on the fixed action space of the \gls{rl} agent, runs in linear time with respect to the number of reward components $N$, and yields optimal subsets at each level of the hierarchy, enabling direct use in agent training.

Algorithm~\ref{alg:precedence_filter} evaluates action relations while preserving the reward preorder. We apply a topological ordering based on the reward precedence~\cite{halder2025sampling}, so that the parents of each reward $r_i$ (i.e., directly connected higher-priority rewards) are always evaluated before $r_i$. At each step, dominance relations established at parent objectives are first inherited: if an action pair $(a,a')$ is already determined to be dominated, dominating, or incomparable, this relation cannot be overridden by a lower-priority objective. In addition, we inherit an action optimal subset $\mathcal{S}^{\uparrow}$ via $\operatorname{Agg}(\cdot)$, which aggregates parent survivor sets, and removes only actions effectively dominated by a surviving non-conflicting dominator.

Only indifferent (undecided) pairs are passed forward for evaluation, where they are compared using the \gls{qd} operator (\ref{sec:pairwise_compare}), to yield local dominance relations. The inherited and local outcomes are merged to update the global dominance structure, while conflicts are filtered out. Finally, the optimal action subset $\mathcal{S}_{r_i}\subseteq \mathcal{S}^{\uparrow}$ is constructed, containing actions that are not strictly dominated under $r_i$. This stepwise filtering propagates precedence consistently through the preorder while pruning actions that fail higher-priority objectives. A key property of the algorithm is that each optimal subset of actions is guaranteed to be non-empty, ensuring that at least one feasible action remains available at every level of the hierarchy.
\begin{algorithm}[ht]
\small
\begin{minipage}{0.95\columnwidth}
\caption{\textsc{Preorder Action Selection}}
\label{alg:precedence_filter}
    
    \KwIn{action set $A$; objectives $\mathcal{R}$ with preorder $\succeq$; parent map $\mathrm{Pa}(\cdot)$; quantile estimates $
    \{ Z^{r_i}\}_{i=1}^N$; comparator $\textsc{QD}(\cdot)$}
    
    \KwOut{for each $r_i \in \mathcal{R}$: optimal subset $\mathcal{S}_{r_i}$}
    
    \BlankLine
    
    $\mathcal{L} \leftarrow \textsc{TopologicalSort}(\mathcal{R}, \succeq)$
    
    \For{$r_i \in \mathcal{L}$}{
      \textbf{\textcolor{blue}{(1) Inherit parent relations}}
      
      \uIf{$\mathrm{Pa}(r_i)=\varnothing$}{
        $\mathcal{S}^{\uparrow}\gets A$;\quad $Dom^{\uparrow}\!\gets\!0 $;\quad $DomBy^{\uparrow}\!\gets\!0$
      }\Else{
        $\mathcal{S}^{\uparrow} \gets \operatorname{Agg}(\{\mathcal{S}_p\}_{p\in \mathrm{Pa}(r_i)})$

        $Dom^{\uparrow}\!\gets\!\bigvee_{p\in\mathrm{Pa}(r_i)} Dom[p]$
        
        $DomBy^{\uparrow}\!\gets\!\bigvee_{p\in\mathrm{Pa}(r_i)} DomBy[p]$
      }
    
      \textcolor{blue}{\textbf{(2) Construct update mask } (only update indifferent pairs)}
      
      $Mask \gets \neg(Dom^{\uparrow} \lor DomBy^{\uparrow})$

      \textcolor{blue}{\textbf{(3) Compare action pairs using QD}}
      
      $(Dom^{\text{QD}}, DomBy^{\text{QD}}) \gets \textsc{QD}(Z^{r_i}, Mask)$
    
      \textbf{\textcolor{blue}{(4) Merge inherited and local}}
      
      $Dom[r_i] \gets (\neg Mask\land Dom^{\uparrow}) \ \lor\ (Mask \land Dom^{\text{QD}})$
      
      $DomBy[r_i] \gets (\neg Mask \land DomBy^{\uparrow}) \ \lor\ (Mask \land DomBy^{\text{QD}})$
    
      \textcolor{blue}{\textbf{(5) Compute optimal subset at reward $r_i$}}
      
      $C\gets Dom[r_i]\land DomBy[r_i]$
      
      $DomBy^{\,\downarrow} \gets DomBy[r_i]\land \neg C$
      
    $\mathcal{S}_{r_i} \gets \{\,a \in \mathcal{S}^{\uparrow} \mid \nexists \, a' \in \mathcal{S}^{\uparrow}:\ DomBy^{\,\downarrow}[a,a']=1\,\}$
    }
    
    \Return $\{\mathcal{S}_{r_i}\}_{r_i\in\mathcal{R}}$
    
    \vspace{0.3ex}
    \hrule height0.2pt
    \vspace{0.3ex}
    
    {\footnotesize
    \textbf{Legend:}
    $\operatorname{Agg}(\cdot)$: aggregation of parent survivor sets;\ 
    $Dom[a,a']\!=\!1$ if $a$ dominates $a'$;\ 
    $DomBy[a,a']\!=\!1$ if $a$ is dominated by $a'$;\
    $\bigvee$ = element-wise logical OR (over parent relations);\
    $Mask$ = undecided action pairs mask;\ 
    $C$ = Incomparable action pairs\\
    \textbf{superscripts:} $\uparrow$ = inherited from parents,\ 
    $\text{QD}$ = computed at $r_i$ via quantile dominance,\ 
    $\downarrow$ = dominated-by after incomparable removal\ 
    }
\end{minipage}
\end{algorithm}
\subsubsection{Preorder Informed Training and Inference} 
Preorder relations between reward components induce optimal action subsets at the value-function level. To leverage these sets during learning, we extend \gls{iqn}~\cite{dabney2018implicit} and denote the resulting algorithm as \emph{Pr-IQN}. Conventional \gls{morl} approaches~\cite{deshpande2021navigation,li2018urban} perform argmax-based target selection over the full action set for each objective, ignoring precedence relations among rewards. In contrast, Pr-IQN modifies the \gls{td}~\cite{sutton1988td} training targets to respect reward precedence. Specifically, we restrict the target selection for each objective $r_i$ to the optimal action subset $\mathcal{S}_{r_i}$ obtained from Alg.~\ref{alg:precedence_filter}, as defined in Eq.~\ref{eq:masked_argmax_mean}. This masking prevents selecting actions that achieve high return for $r_i$ while violating higher-priority objectives. Accordingly, the \gls{td} error between quantiles $(\tau,\tau')$, denoted $\delta^{r_i,(\tau,\tau')}_t$, is computed as in Eq.~\ref{eq:td_error}, ensuring that value updates promote actions that optimize the current objective while respecting precedence constraints.
\begin{align}
  a^{*,r_i}_{t+1}
  &\;=\;
  \operatorname*{arg\,max}_{a \in \mathcal{S}_{r_i}}
  \frac{1}{|\tau|}\sum_{\tau_k \in \tau} Z^{r_i}_{\tau_k}(o_{t+1},a)
  \label{eq:masked_argmax_mean}\\
\delta^{r_i, \,(\tau, \tau')}_{t} &= r^{\,i}_t + \gamma\, Z^{r_i}_{\tau'}(o_{t+1}, a^{*,r_i}_{t+1}) -  Z^{r_i}_{\tau}(o_{t}, a_{t})
  \label{eq:td_error}
\end{align}
During inference, the agent samples an action uniformly from the optimal subset associated with a leaf objective, following the approach in~\cite{halder2025sampling}. 
When the hierarchy contains multiple leaves, we introduce a virtual global leaf that aggregates their optimal subsets and guides action selection. 

\section{experimental Setup}
\label{sec:experiments}
This section details the experimental setup, including the RL agent design, hierarchical reward structure, and urban traffic scenarios in CARLA~\cite{dosovitskiy2017carla}. We also outline baselines, ablations, and evaluation metrics to enable a systematic and fair comparison of performance.
\subsection{RL Agent Description}
We design a multimodal observation space that combines a front-facing RGB camera with resolution $128 \times 128$ and a LiDAR point cloud projected onto a $128 \times 128$ grid map with two vertical bins. Additionally, the agent is conditioned on high-level navigational commands~\cite{chitta2022transfuser} and on vehicle kinematics, including longitudinal and lateral velocities and accelerations. 
To encode this observation, we employ TransFuser~\cite{chitta2022transfuser}, a transformer-based backbone that fuses image and LiDAR features into a shared latent representation.

For decision-making, we couple the \gls{rl} agent with a Frenet-based planner~\cite{bogdoll2024informed}, which generates trajectories consistent with road geometry. The agent outputs two discrete boundary conditions $(v_f, d_f)$: $v_f$ denotes the target velocity at the end of the planning horizon, and $d_f$ the lateral displacement from the lane centerline. These conditions are used by the Frenet planner to construct a feasible trajectory.
\subsection{Reward Hierarchy}
\label{section:reward_hierarchy}
\begin{figure}[t]
    \centering
    \includegraphics[width=0.9\columnwidth]{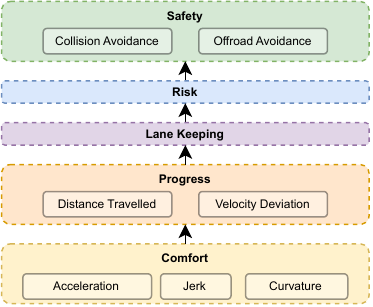}
    \caption[Hierarchy of Reward Terms]{The reward hierarchy with Safety as the highest priority, followed by Risk, Lane Keeping, Progress, and Comfort. This ordering guides the agent’s decision-making to emphasize safety while balancing other objectives.} 
    \label{fig:reward_hierarchy} 
\end{figure}

The reward hierarchy illustrated in Fig.~\ref{fig:reward_hierarchy} organizes driving objectives according to their criticality for safe and reliable \gls{ad}. \emph{Safety} has the highest priority, as collision and off-road events are enforced as first-order constraints due to their catastrophic consequences~\cite{censi2019liability}. The second level addresses \emph{risk mitigation}, encouraging conservative driving behavior by maintaining clearance and proactively reducing collision likelihood~\cite{Balancing_Abouelazm_2025}. Placing risk directly below safety ensures that near-miss situations are penalized before progress incentives can dominate. Below risk, \emph{lane keeping} enforces compliance with road geometry, supporting both safety and predictability in mixed-traffic~\cite{helou2021reasonable}. \emph{Progress} follows, rewarding efficient route advancement and adherence to target velocity profiles. Finally, \emph{comfort} is assigned the lowest priority as it primarily affects ride quality rather than immediate safety.
\subsection{Traffic Scenarios} 
In this work, we focus on urban driving tasks where an autonomous agent must approach and cross unsignalized intersections. Such intersections are among the most safety-critical elements of road networks due to the absence of explicit right-of-way indicators and the need for implicit negotiation with other vehicles~\cite{al2024autonomous}. While our framework applies to a broad range of road scenarios, intersections provide a particularly demanding setting for evaluating risk-sensitive \gls{rl} strategies.

Traffic scenarios are generated in CARLA~\cite{dosovitskiy2017carla}, where training involves randomized configurations of static obstacles and traffic vehicles across multiple T-junctions and four-way intersections. Vehicle attributes such as geometry, speed, and lateral positioning are randomized to promote robustness and generalization. For evaluation, we adopt a hold-out set consisting of one unseen T-junction and two unseen four-way intersections, ensuring that performance is assessed on layouts not encountered during training.
\subsection{Baselines and Evaluation Metrics} 
We benchmark our \gls{rl} framework against \gls{iqn}~\cite{dabney2018implicit}, a widely used distributional \gls{rl} baseline. To control for model capacity, we introduce an ensemble IQN variant with the same number of policy heads as our reward hierarchy, allowing us to disentangle gains from increased capacity and those from explicitly encoding semantic structure. Both baselines are trained using a weighted sum of the hierarchical reward components described in Section~\ref{section:reward_hierarchy}, following the weighting scheme of~\cite{Balancing_Abouelazm_2025}.

Additionally, we adopt the multi-objective approach of~\cite{juozapaitis2019explainable,Seijen_hybrid_reward_17}, which learns one value head per objective $r_i$ and selects actions using the mean value across objectives, denoted as \gls{maiqn}. To isolate the contribution of each component of our framework, we conduct ablation studies examining the effect of allocating separate policy heads per objective, integrating hierarchical comparisons during training, and varying both the comparison method and threshold. We also evaluate the method under partial ordering, using a hierarchy in which risk and lane keeping are children of safety and precede progress.

To ensure a fair comparison, all agents are trained for the same number of steps using identical architectures and hyperparameters. Training is repeated with three random seeds, and each policy is evaluated over three runs to account for stochasticity in CARLA, following the protocol of~\cite{jaeger2025carl}. Evaluation uses a hold-out set of intersection scenarios with varying traffic densities, defined as the ratio of active actors to the maximum allowed in the environment.

We evaluate performance using driving metrics and statistical reliability measures. Driving performance is measured using success rate ($SR$), off-road rate ($OR$), collision rate ($CR$), and route progress ($RP$), reported as mean $\pm$ standard deviation across all seeds and runs. We also assess the agent’s ability to optimize individual reward components, reflecting alignment with the designed objectives. To complement these metrics, we use the RLiable library~\cite{agarwal2021deep} to compute statistics such as the interquartile mean (IQM) and optimality gap.
\section{Evaluation}
\begin{figure}[t]
    \centering
    \includegraphics[width=0.85\linewidth,trim=8 8 8 0,clip]{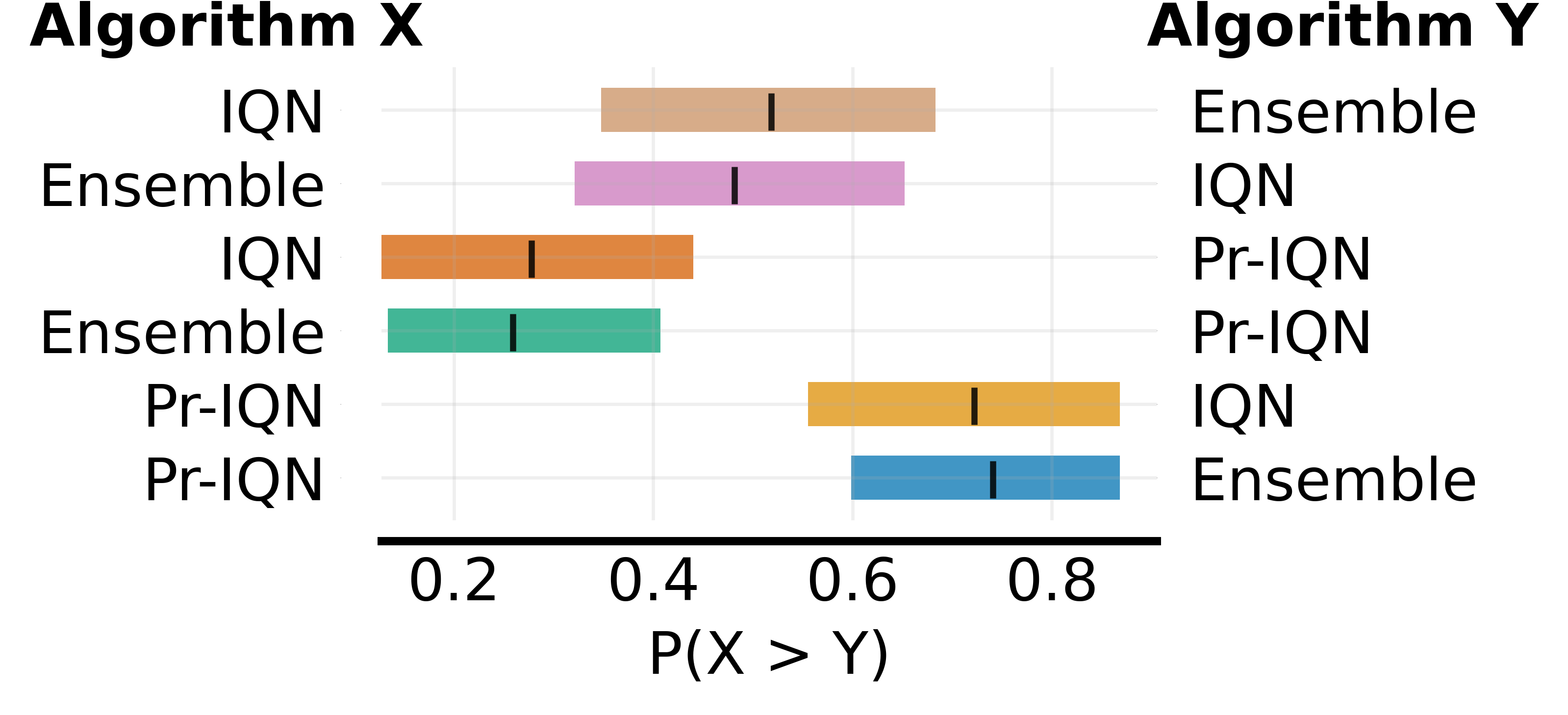}
    \caption{Probability of improvement~\cite{agarwal2021deep}, quantifying the likelihood that an algorithm X (the left column) outperforms algorithm Y (the right column).}
    \label{fig:prob}
\end{figure}
Table~\ref{tab:traffic_density_grouped} reports an ablation study analyzing the impact of the comparison metrics, the tolerance $\epsilon$, and the integration of the preorder during training. MA-IQN performs noticeably worse since averaging value estimates across heads collapses the hierarchical ordering, allowing lower-priority improvements to compensate for violations of higher-priority objectives. Similarly, collapsing quantile estimates into scalar metrics such as \gls{cvar} or \gls{mv} in Pr-IQN removes the distributional structure, leading to unreliable optimal action subsets.

Pr-IQN with QD consistently outperforms IQN and Ensemble across all traffic densities. Tightening the tolerance from $\epsilon=0.4$ to $\epsilon=0.2$ yields further gains by enforcing stricter adherence to the preorder. Incorporating the preorder during training further improves the success rate by $+3.3\%$ and $+2.5\%$ at densities $0.75$ and $1.0$. A partial preorder performs comparably to the total-order variant when the hierarchy is enforced during training, but is more sensitive to degradation when it is not, highlighting the importance of aligning the training objective with the decision structure.

Overall, our best configuration, $\text{Pr-IQN}^*$ (QD with total preorder enforced during training and $\epsilon=0.2$), improves success rate by $(+7.7\%, +16.6\%, +20.3\%)$ over IQN and $(+11.0\%, +14.7\%, +13.9\%)$ over Ensemble-IQN at traffic densities 0.5, 0.75, and 1.0. These results show that preorder-guided optimal subsets prevent policies from exploiting lower-priority objectives at the expense of safety.

Additionally, Table~\ref{tab:traffic_density_rewards} compares policies' ability to optimize the top three reward components in the preorder: safety, risk, and lane-keeping. The table reports the mean and standard deviation of cumulative rewards per episode, along with the relative percentage improvement over IQN. Results show that $\text{Pr-IQN}^*$ consistently increases rewards across all traffic densities, achieving improvements of up to $61\%$ in safety violations, $41\%$ in risk exposure, and $37\%$ in lane-keeping rewards. These improvements highlight that explicitly incorporating the preorder not only enhances overall task performance but also yields safer and more reliable driving behavior by directly prioritizing high-criticality objectives.

\begin{figure}[t]
    \centering
    \includegraphics[width=1.0\linewidth,trim=8 8 8 2,clip]{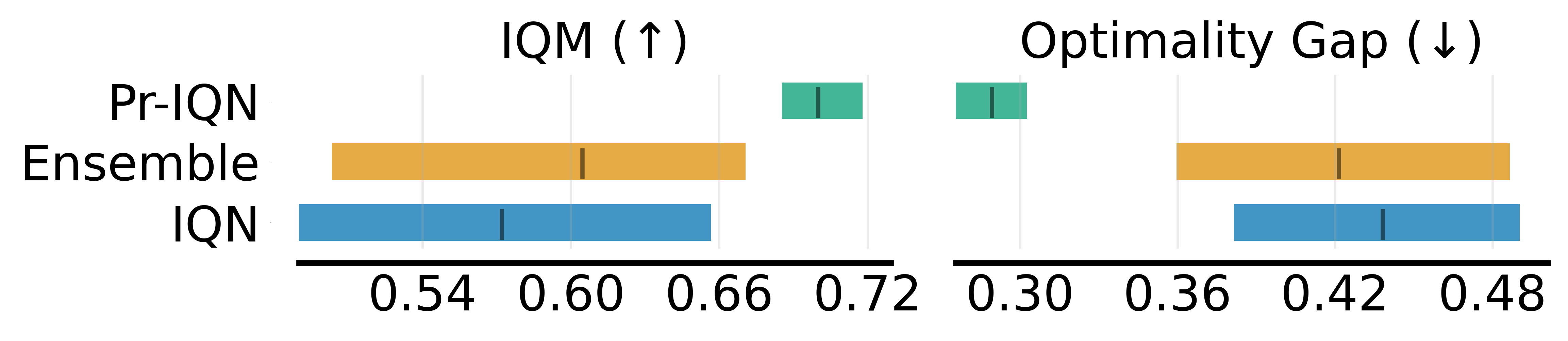}
    \caption{Interquartile mean (IQM) and optimality gap~\cite{agarwal2021deep}, quantifying the statistical stability of a policy.}
    \label{fig:iqm}
\end{figure}

Figures~\ref{fig:prob} and~\ref{fig:iqm} complement these findings using RLiable metrics across training seeds and evaluation runs. $\text{Pr-IQN}^*$ consistently achieves the highest IQM and lowest optimality gap, confirming its reliability over IQN and Ensemble-IQN. The probability of improvement analysis further shows $P(\text{Pr-IQN} > \text{IQN})$ and $P(\text{Pr-IQN} > \text{Ensemble-IQN})$ substantially above $0.5$, while the reverse probabilities remain low. These results highlight that incorporating preorder relations into distributional \gls{rl} improves not only average performance but also stability and robustness across runs.

\section{Conclusion}
We introduced Pr-MOMDP to encode reward preorder, proposed Quantile Dominance (QD) for distribution-aware action comparison, and developed an algorithm to extract optimal action subsets consistent with the preorder. Leveraging these, we extended IQN into Pr-IQN, where optimal subsets shape both training and decision-making. Experiments in CARLA show that Pr-IQN improves safety, success rate, and overall driving performance compared to IQN and ensemble baselines, with gains of up to 7.7\%--20.3\% over IQN and 11.0\%--14.7\% over Ensemble-IQN across traffic densities. Future work will address scalability to larger preorders and investigate how the multi-head architecture can be used for explainability and targeted fine-tuning of specific objectives. We also plan to explore hybrid setups, where certain objectives, e.g., traffic-light compliance, are evaluated by external components such as Car2X and integrated into the preorder.

\section*{ACKNOWLEDGMENT}
The research leading to these results is funded by the German Federal Ministry for Economic Affairs and Energy within the project “Safe AI Engineering – Sicherheitsargumentation befähigendes AI Engineering über den gesamten Lebenszyklus einer KI-Funktion". The authors would like to thank the consortium for the successful cooperation.
\begin{table}[t]
\centering 
\caption{Rewards for multiple objectives of different policies across traffic densities. $\Delta(\%)$ denotes the relative reward gain expressed as a percentage with respect to IQN. Higher values indicate better alignment of the policy with the defined objectives.}
\label{tab:traffic_density_rewards} 
\setlength{\tabcolsep}{3pt} 
\renewcommand{\arraystretch}{1.2} 
\resizebox{\columnwidth}{!}{%
\begin{tabular}{P{0.95cm} c c c c c c} 
\toprule 
\multirow{2}{*}{\centering Policy} &
\multicolumn{6}{c}{\textbf{Reward components}} \\
\cmidrule(lr){2-7}
& Safety$\uparrow$ & $\Delta$ (\%) & Risk$\uparrow$ & $\Delta$ (\%) & Lane keeping$\uparrow$ & $\Delta$ (\%) \\
\midrule 
\multicolumn{7}{c}{\textbf{Traffic Density 0.5}} \\
\midrule 
IQN & $-0.007\pm0.003$ & -- & $-0.086\pm0.016$ & -- & $-0.120\pm0.013$ & -- \\
Ensemble & $-0.008\pm0.006$ & $-9.4$ & $-0.104\pm0.026$ & $-20.8$ & $-0.124\pm0.065$ & $-3.0$ \\

$\text{Pr-IQN}^*$ & $\bm{-0.004\pm0.002}$ & $\bm{+42.0}$ & $\bm{-0.051\pm0.011}$ & $\bm{+40.9}$ & $\bm{-0.076\pm0.013}$ & $\bm{+36.8}$ \\
\midrule

\multicolumn{7}{c}{\textbf{Traffic Density 0.75}} \\
\midrule
IQN & $-0.012\pm0.006$ & -- & $-0.137\pm0.0281$ & -- & $-0.126\pm0.017$ & -- \\
Ensemble & $-0.011\pm0.010$ & $+7.2$ & $-0.168\pm0.042$ & $-22.8$ & $-0.110\pm0.041$ & $+12.6$ \\
$\text{Pr-IQN}^*$ & $\bm{-0.006\pm0.001}$ & $\bm{+51.5}$ & $\bm{-0.087\pm0.014}$ & $\bm{+36.4}$ & $\bm{-0.080\pm0.011}$ & $\bm{+36.2}$ \\
\midrule

\multicolumn{7}{c}{\textbf{Traffic Density 1.0}} \\
\midrule
IQN & $-0.017\pm0.009$ & -- & $-0.164\pm0.026$ & -- & $-0.130\pm0.021$ & -- \\
Ensemble & $-0.013\pm0.010$ & $+21.7$ & $-0.200\pm0.038$ & $-21.4$ & $-0.110\pm0.042$ & $+15.2$ \\

$\text{Pr-IQN}^*$ & $\bm{-0.006\pm0.001}$ & $\bm{+61.1}$ & $\bm{-0.104\pm0.014}$ & $\bm{+36.6}$ & $\bm{-0.092\pm0.006}$ & $\bm{+29.1}$ \\ \bottomrule 
\end{tabular}
}
\end{table}
\begin{table*}[t]
\centering
\caption{Evaluation metrics of different policies with various comparison metrics and thresholds across various traffic densities.}
\label{tab:traffic_density_grouped}
\setlength{\tabcolsep}{4pt}
\renewcommand{\arraystretch}{1.2}
\begin{tabular}{P{1.3cm} P{1.3cm} P{1.3cm} P{1.3cm} P{1.3cm} c c c c}
\toprule
\multirow{2}{1.3cm}{\centering Policy} &
\multirow{2}{1.3cm}{\centering Comparison Metric} &
\multirow{2}{1.3cm}{\centering Training Preorder} &
\multirow{2}{1.3cm}{\centering Threshold $\boldsymbol{\epsilon}$} &
\multirow{2}{1.3cm}{\centering Partial Order} &
\multicolumn{4}{c}{\textbf{Evaluation metrics}} \\
\cmidrule(lr){6-9}
& & & & & CR $\downarrow$ & OR $\downarrow$ & SR $\uparrow$ & RP $\uparrow$ \\
\midrule

\multicolumn{9}{c}{\textbf{Traffic Density 0.5}} \\
\midrule
IQN   & --     & --     & --   & -- & $0.248 \pm 0.082$ & $0.013 \pm 0.015$ & $0.737 \pm 0.084$ & $0.764 \pm 0.053$ \\
Ensemble & --     & --     & --   & -- & $0.272 \pm 0.145$  & $0.023 \pm 0.022$ & $0.704 \pm  0.160$ & $ 0.782 \pm 0.065$ \\
MA-IQN & --     & --     & --   & -- & $0.403 \pm 0.039$  & $0.002 \pm 0.004$ & $0.594 \pm  0.042$ & $ 0.696 \pm 0.027$ \\

Pr-IQN & \gls{mv} & \ding{51} & $0.4$ & \ding{55} & $0.333\pm0.107$ & $0.031\pm0.015$ & $0.635\pm 0.106$ & $0.712\pm0.052$ \\
Pr-IQN & \gls{cvar}   & \ding{51} & $0.4$ & \ding{55} & $0.448\pm0.057$ & $0.032\pm0.023$ & $0.518\pm0.066$ & $0.641\pm0.054$ \\

Pr-IQN & \gls{qd}   & \ding{55} & $0.4$ & \ding{55} & $0.245\pm 0.050$ & $0.032\pm 0.014$ & $0.722\pm 0.047$ & $0.767\pm 0.035$ \\
Pr-IQN & \gls{qd}   & \ding{51} & $0.4$ & \ding{55} & $0.217\pm0.023$ & $0.018\pm0.012$ & $0.763\pm0.029$ & $0.789\pm0.020$ \\
Pr-IQN & \gls{qd}   & \ding{55} & $0.2$ & \ding{55} & $\bm{0.136\pm0.041}$ & $\underline{0.022\pm0.011}$ & $\underline{0.816\pm0.047}$ & $\bm{0.830\pm 0.028}$ \\
Pr-IQN & \gls{qd}   & \ding{51} & $0.2$ & \ding{55} & $0.175\pm0.029$ & $\bm{0.010\pm0.011}$ & 
$\bm{0.816\pm0.030}$ & $\underline{0.808\pm0.019}$ \\

Pr-IQN & \gls{qd}   & \ding{55} & $0.2$ & \ding{51} & $ 0.250 \pm 0.087$ & $ 0.034 \pm 0.018$ & $ 0.692 \pm  0.087 $ & $0.766 \pm 0.055$ \\

Pr-IQN & \gls{qd}   & \ding{51} & $0.2$ & \ding{51} & $ \underline{0.161 \pm 0.046}$ & $0.032 \pm 0.035 $ & $ 0.797\pm  0.087$ & $0.805 \pm 0.058$ \\
\midrule

\multicolumn{9}{c}{\textbf{Traffic Density 0.75}} \\
\midrule
IQN   & --     & --     & --   & -- & $0.472 \pm 0.149$ & $0.014 \pm 0.011$ & $0.513\pm 0.156$ & $0.611 \pm 0.105$ \\
Ensemble & --     & --     & --   & -- & $0.451 \pm 0.171$ & $0.016\pm0.017$ & $0.532\pm 0.184$ & $0.670\pm0.099$ \\
MA-IQN & --     & --     & --   & -- & $0.617 \pm 0.063$  & $0.008 \pm 0.009$ & $0.374 \pm  0.065$ & $ 0.515 \pm 0.045$ \\

Pr-IQN & \gls{mv} & \ding{51} & $0.4$ & \ding{55} & $0.570\pm0.142$ & $0.030\pm0.014$ & $0.400\pm0.145$ & $0.536\pm0.092$ \\
Pr-IQN & \gls{cvar}   & \ding{51} & $0.4$ & \ding{55} & $0.692\pm0.062$ & $0.032\pm0.024$ & $0.275\pm 0.079$ & $0.451\pm0.073$ \\

Pr-IQN & \gls{qd}   & \ding{55} & $0.4$ & \ding{55} & $0.416\pm 0.058$ & $0.015\pm0.011$ & $0.567\pm 0.057$ & $0.661\pm0.039$ \\
Pr-IQN & \gls{qd}   & \ding{51} & $0.4$ & \ding{55} & $0.403\pm0.055$ & $0.016\pm0.012$ & $0.580\pm0.062$ & $0.673\pm0.049$ \\
Pr-IQN & \gls{qd}   & \ding{55} & $0.2$ & \ding{55} & $\bm{0.277\pm0.023}$ & $\underline{0.032\pm0.030}$ & $\underline{0.646\pm0.030}$ & $\underline{0.713\pm0.020}$ \\
Pr-IQN & \gls{qd}   & \ding{51} & $0.2$ & \ding{55} & $\underline{0.314\pm0.030}$ & $\bm{0.005\pm0.004}$ & $\bm{0.679\pm0.032}$ & $\bm{0.717\pm0.026}$ \\

Pr-IQN & \gls{qd}   & \ding{55} & $0.2$ & \ding{51} & $ 0.440\pm 0.138$ & $ 0.028\pm 0.018$ & $ 0.514 \pm 0.168$ & $ 0.643 \pm 0.088$ \\
Pr-IQN & \gls{qd}   & \ding{51} & $0.2$ & \ding{51} & $ 0.337 \pm 0.050$ & $ 0.027 \pm 0.023$ & $0.630 \pm 0.072$ & $ 0.682 \pm $ 0.051\\

\midrule

\multicolumn{9}{c}{\textbf{Traffic Density 1.0}} \\
\midrule
IQN   & --     & --     & --   & -- & $0.538 \pm 0.163$ & $0.026 \pm 0.034$ & $0.434 \pm 0.177$ & $0.558 \pm 0.129$ \\
Ensemble & --     & --     & --   & -- & $0.485\pm0.156$ & $0.015\pm0.017$ & $0.498\pm0.164$& $0.652\pm0.091$ \\
MA-IQN & --     & --     & --   & -- & $0.604 \pm 0.045$  & $0.004 \pm 0.005$  & $ 0.391 \pm 0.048$ & $0.532 \pm  0.037$ \\

Pr-IQN & \gls{mv} & \ding{51} & $0.4$ & \ding{55} & $0.616\pm0.124$ & $0.017\pm0.015$ & $0.365\pm0.118$ & $0.500\pm 0.075$ \\
Pr-IQN & \gls{cvar}   & \ding{51} & $0.4$ & \ding{55} & $0.738\pm0.052$ & $0.036\pm0.036$ & $0.224\pm0.063$ & $0.413\pm0.071$ \\

Pr-IQN & \gls{qd}   & \ding{55} & $0.4$ & \ding{55} & $0.507\pm 0.060$ & $0.028\pm0.018$ & $0.463\pm0.054$ & $0.594\pm0.045$ \\
Pr-IQN & \gls{qd}   & \ding{51} & $0.4$ & \ding{55} & $0.512\pm0.044$ & $0.032\pm0.018$ & $0.455\pm0.054$ & $0.590\pm0.047$ \\
Pr-IQN & \gls{qd}   & \ding{55} & $0.2$ & \ding{55} & $\bm{0.335\pm0.045}$ & $\underline{0.013\pm0.006}$ & $0.612\pm0.044$ & $\underline{0.693\pm0.028}$ \\
Pr-IQN & \gls{qd}   & \ding{51} & $0.2$ & \ding{55} & $0.353\pm0.045$ & $\bm{0.007\pm0.010}$ & $\underline{0.637\pm0.043}$ & $0.688\pm0.039$ \\

Pr-IQN & \gls{qd}   & \ding{55} & $0.2$ & \ding{51} & $ 0.453\pm 0.125$ & $ 0.034\pm 0.015$ & $0.503 \pm 0.141$ & $ 0.634 \pm 0.075$ \\
Pr-IQN & \gls{qd}   & \ding{51} & $0.2$ & \ding{51} & $ \underline{0.343 \pm 0.036} $ & $0.014 \pm 0.005$ & $ \bm{0.642 \pm 0.035} $ & $\bm{0.697 \pm 0.032}$ \\

\bottomrule

\end{tabular}
\end{table*}


{
    \bibliographystyle{IEEEtran}
    \bibliography{references}

@IEEEtranBSTCTL{IEEEexample:BSTcontrol,
CTLuse_forced_etal       = "yes",
CTLmax_names_forced_etal = "3",
CTLnames_show_etal       = "1" }

@article{hanAutoRewardClosedLoopReward2024,
  title = {{{AutoReward}}: {{Closed-Loop Reward Design}} with {{Large Language Models}} for {{Autonomous Driving}}},
  author = {Han, Xu and Yang, Qiannan and Chen, Xianda and Cai, Zhenghan and Chu, Xiaowen and Zhu, Meixin},
  year = {2024},
  journal = {IEEE Transactions on Intelligent Vehicles}
}

@inproceedings{macglashanValueFunctionDecomposition,
  title = {Value {{Function Decomposition}} for {{Iterative Design}} of {{Reinforcement Learning Agents}}},
  author = {MacGlashan, James and Archer, Evan and Devlic, Alisa and Seno, Takuma and Sherstan, Craig and Wurman, Peter R and Stone, Peter},
year = {2025},
booktitle = {NeurIPS}
}

@inproceedings{alegreAMORAdaptiveCharacter2025,
  title = {{{AMOR}}: {{Adaptive Character Control}} through {{Multi-Objective Reinforcement Learning}}},
  booktitle = {{{Special Interest Group}} on {{Computer Graphics}} and {{Interactive Techniques}}},
  author = {Alegre, Lucas N. and Serifi, Agon and Grandia, Ruben and M{\"u}ller, David and Knoop, Espen and B{\"a}cher, Moritz},
  year = {2025},
}

@article{jamilDynamicWeightbasedMultiObjective2022,
  title = {Dynamic {{Weight-based Multi-Objective Reward Architecture}} for {{Adaptive Traffic Signal Control System}}},
  author = {Jamil, Abu Rafe Md and Nower, Naushin},
  year = {2022},
  journal = {International Journal of Intelligent Transportation Systems Research},
}

@inproceedings{juozapaitis2019explainable,
  title={Explainable reinforcement learning via reward decomposition},
  author={Juozapaitis, Zoe and Koul, Anurag and Fern, Alan and Erwig, Martin and Doshi-Velez, Finale},
  booktitle={IJCAI/ECAI Workshop on explainable artificial intelligence},
  year={2019}
}

@INPROCEEDINGS{joseph_dreamtodrive_2024,
  author={Joseph, Tim and Fechner, Marcus and Abouelazm, Ahmed and Zöllner, Marius J.},
  booktitle={IEEE International Conference on Intelligent Transportation Systems (ITSC)}, 
  title={Dream to Drive: Learning Conditional Driving Policies in Imagination}, 
  year={2024},
}

@article{surmann2025multiobjectivereinforcementlearningadaptable,
      title={Multi-Objective Reinforcement Learning for Adaptable Personalized Autonomous Driving}, 
      author={Hendrik Surmann and Jorge de Heuvel and Maren Bennewitz},
      year={2025},
      journal={arXiv:2505.05223},
}

@inproceedings{skalseDefiningCharacterizingReward2022,
  title = {Defining and Characterizing Reward Hacking},
  booktitle = {{{International Conference}} on {{Neural Information Processing Systems}}},
  author = {Skalse, Joar and Howe, Nikolaus H. R. and Krasheninnikov, Dmitrii and Krueger, David},
  year = {2022},
}

@article{endtoendreview,
  title={A review of end-to-end autonomous driving in urban environments},
  author={Coelho, Daniel and Oliveira, Miguel},
  journal={IEEE Access},
  year={2022},
}

@inproceedings{hu2023planning,
  title={Planning-oriented autonomous driving},
  author={Hu, Yihan and Yang, Jiazhi and Chen, Li and Li, Keyu and Sima, Chonghao and Zhu, Xizhou and Chai, Siqi and Du, Senyao and Lin, Tianwei and Wang, Wenhai and others},
  booktitle={Conference on computer vision and pattern recognition},
  year={2023}
}

@article{tampuu2020survey,
  title={A survey of end-to-end driving: Architectures and training methods},
  author={Tampuu, Ardi and Matiisen, Tambet and Semikin, Maksym and Fishman, Dmytro and Muhammad, Naveed},
  journal={IEEE Transactions on Neural Networks and Learning Systems},
  year={2020},
}

@article{kiran2021deep,
  title={Deep reinforcement learning for autonomous driving: A survey},
  author={Kiran, B Ravi and Sobh, Ibrahim and Talpaert, Victor and Mannion, Patrick and Al Sallab, Ahmad A and Yogamani, Senthil and P{\'e}rez, Patrick},
  journal={IEEE transactions on intelligent transportation systems},
  year={2021},
}

@article{chen2024end,
  title={End-to-end autonomous driving: Challenges and frontiers},
  author={Chen, Li and Wu, Penghao and Chitta, Kashyap and Jaeger, Bernhard and Geiger, Andreas and Li, Hongyang},
  journal={IEEE Transactions on Pattern Analysis and Machine Intelligence},
  year={2024},
}

@inproceedings{abouelazm2024review,
  title={A review of reward functions for reinforcement learning in the context of autonomous driving},
  author={Abouelazm, Ahmed and Michel, Jonas and Z{\"o}llner, J Marius},
  booktitle={IEEE Intelligent Vehicles Symposium (IV)},
  year={2024},
}

@article{knox2023reward,
  title={Reward (mis) design for autonomous driving},
  author={Knox, W Bradley and Allievi, Alessandro and Banzhaf, Holger and Schmitt, Felix and Stone, Peter},
  journal={Artificial Intelligence},
  year={2023},
}

@inproceedings{zhangDistributionalReinforcementLearning2021,
  title = {Distributional {{Reinforcement Learning}} for {{Multi-Dimensional Reward Functions}}},
  booktitle = {NeurIPS},
  author = {Zhang, Pushi and Chen, Xiaoyu and Zhao, Li and Xiong, Wei and Qin, Tao and Liu, Tie-Yan},
  year = {2021},
}

@article{jinHybridActionBased2025,
  title = {Hybrid {{Action Based Reinforcement Learning}} for {{Multi-Objective Compatible Autonomous Driving}}},
  author = {Jin, Guizhe and Li, Zhuoren and Leng, Bo and Han, Wei and Xiong, Lu and Sun, Chen},
  year = {2025},
  journal = {arXiv:2501.08096}
}

@inproceedings{wiltzerFoundationsMultivariateDistributional,
  title = {Foundations of {{Multivariate Distributional Reinforcement Learning}}},
  author = {Wiltzer, Harley and Farebrother, Jesse and Gretton, Arthur and Rowland, Mark},
year = {2024},
booktitle = {NeurIPS}}

@inproceedings{deshpande2021navigation,
  title={Navigation in urban environments amongst pedestrians using multi-objective deep reinforcement learning},
  author={Deshpande, Niranjan and Vaufreydaz, Dominique and Spalanzani, Anne},
  booktitle={IEEE International Intelligent Transportation Systems Conference (ITSC)},
  year={2021},
}

@inproceedings{bogdoll2024informed,
  title={Informed reinforcement learning for situation-aware traffic rule exceptions},
  author={Bogdoll, Daniel and Qin, Jing and Nekolla, Moritz and Abouelazm, Ahmed and Joseph, Tim and Z{\"o}llner, J Marius},
  booktitle={2024 IEEE International Conference on Robotics and Automation (ICRA)},
  year={2024},
}

@INPROCEEDINGS{Balancing_Abouelazm_2025,
  author={Abouelazm, Ahmed and Michel, Jonas and Gremmelmaier, Helen and Joseph, Tim and Schörner, Philip and Zöllner, J. Marius},
  booktitle={2025 IEEE Intelligent Vehicles Symposium (IV)}, 
  title={Balancing Progress and Safety: A Novel Risk-Aware Objective for RL in Autonomous Driving}, 
  year={2025},
}

@inproceedings{censi2019liability,
  title={Liability, ethics, and culture-aware behavior specification using rulebooks},
  author={Censi, Andrea and Slutsky, Konstantin and Wongpiromsarn, Tichakorn and Yershov, Dmitry and Pendleton, Scott and Fu, James and Frazzoli, Emilio},
  booktitle={2019 international conference on robotics and automation (ICRA)},
  year={2019},
}

@inproceedings{yuan2019multi,
  title={Multi-reward architecture based reinforcement learning for highway driving policies},
  author={Yuan, Wei and Yang, Ming and He, Yuesheng and Wang, Chunxiang and Wang, Bing},
  booktitle={IEEE Intelligent Transportation Systems Conference (ITSC)},
  year={2019},
}

@article{chitta2022transfuser,
  title={Transfuser: Imitation with transformer-based sensor fusion for autonomous driving},
  author={Chitta, Kashyap and Prakash, Aditya and Jaeger, Bernhard and Yu, Zehao and Renz, Katrin and Geiger, Andreas},
  journal={IEEE Transactions on Pattern Analysis and Machine Intelligence},
  year={2022},
}

@inproceedings{Seijen_hybrid_reward_17,
author = {van Seijen, Harm and Fatemi, Mehdi and Romoff, Joshua and Laroche, Romain and Barnes, Tavian and Tsang, Jeffrey},
title = {Hybrid reward architecture for reinforcement learning},
year = {2017},
booktitle = {NeurIPS},
}

@INPROCEEDINGS{yuan_multireward_2019,
  author={Yuan, Wei and Yang, Ming and He, Yuesheng and Wang, Chunxiang and Wang, Bing},
  booktitle={IEEE Intelligent Transportation Systems Conference (ITSC)}, 
  title={Multi-Reward Architecture based Reinforcement Learning for Highway Driving Policies}, 
  year={2019},
}

@inproceedings{liUrbanDrivingMultiObjective2019,
  title = {Urban {{Driving}} with {{Multi-Objective Deep Reinforcement Learning}}},
  author = {Li, Changjian and Czarnecki, Krzysztof},
  year = {2019},
  booktitle = {AAMAS},
}

@article{al2024autonomous,
  title={Autonomous driving at unsignalized intersections: A review of decision-making challenges and reinforcement learning-based solutions},
  author={Al-Sharman, Mohammad and Edes, Luc and Sun, Bert and Jayakumar, Vishal and Daoud, Mohamed A and Rayside, Derek and Melek, William},
  journal={arXiv:2409.13144},
  year={2024}
}

@inproceedings{dosovitskiy2017carla,
  title={CARLA: An open urban driving simulator},
  author={Dosovitskiy, Alexey and Ros, German and Codevilla, Felipe and Lopez, Antonio and Koltun, Vladlen},
  booktitle={Conference on robot learning},
  year={2017},
}

@inproceedings{linDistributionalRewardDecomposition2019,
  title = {Distributional {{Reward Decomposition}} for {{Reinforcement Learning}}},
  booktitle = {NeurIPS},
  author = {Lin, Zichuan and Zhao, Li and Yang, Derek and Qin, Tao and Liu, Tie-Yan and Yang, Guangwen},
  year = {2019},
}

@inproceedings{caiDistributionalParetoOptimalMultiObjective,
  title = {Distributional {{Pareto-Optimal Multi-Objective Reinforcement Learning}}},
  author = {Cai, Xin-Qiang and Zhang, Pushi and Zhao, Li and Bian, Jiang and Sugiyama, Masashi and Llorens, Ashley J},
year = {2023},
booktitle = {NeurIPS}
}

@inproceedings{dabney2018implicit,
  title={Implicit quantile networks for distributional reinforcement learning},
  author={Dabney, Will and Ostrovski, Georg and Silver, David and Munos, R{\'e}mi},
  booktitle={International conference on machine learning},
  year={2018},
}

@inproceedings{helou2021reasonable,
  title={The reasonable crowd: Towards evidence-based and interpretable models of driving behavior},
  author={Helou, Bassam and Dusi, Aditya and Collin, Anne and Mehdipour, Noushin and Chen, Zhiliang and Lizarazo, Cristhian and Belta, Calin and Wongpiromsarn, Tichakorn and Tebbens, Radboud Duintjer and Beijbom, Oscar},
  booktitle={IEEE/RSJ International Conference on Intelligent Robots and Systems (IROS)},
  year={2021},
}

@article{jaeger2025carl,
  title={Carl: Learning scalable planning policies with simple rewards},
  author={Jaeger, Bernhard and Dauner, Daniel and Bei{\ss}wenger, Jens and Gerstenecker, Simon and Chitta, Kashyap and Geiger, Andreas},
  journal={arXiv:2504.17838},
  year={2025}
}

@inproceedings{halder2025sampling,
  title={Sampling-Based Motion Planning with Preordered Objectives},
  author={Halder, Patrick and Althoff, Matthias},
  booktitle={2025 IEEE Intelligent Vehicles Symposium (IV)},
  year={2025},
}

@article{li2018urban,
  title={Urban driving with multi-objective deep reinforcement learning},
  author={Li, Changjian and Czarnecki, Krzysztof},
  journal={arXiv:1811.08586},
  year={2018}
}

@article{theate2023risk,
  title={Risk-sensitive policy with distributional reinforcement learning},
  author={Th{\'e}ate, Thibaut and Ernst, Damien},
  journal={Algorithms},
  year={2023},
}

@article{agarwal2021deep,
  title={Deep Reinforcement Learning at the Edge of the Statistical Precipice},
  author={Agarwal, Rishabh and Schwarzer, Max and Castro, Pablo Samuel
          and Courville, Aaron and Bellemare, Marc G},
  journal={NeurIPS},
  year={2021}
}

@inproceedings{lu2023imitation,
  title={Imitation is not enough: Robustifying imitation with reinforcement learning for challenging driving scenarios},
  author={Lu, Yiren and Fu, Justin and Tucker, George and Pan, Xinlei and Bronstein, Eli and Roelofs, Rebecca and Sapp, Benjamin and White, Brandyn and Faust, Aleksandra and Whiteson, Shimon and others},
  booktitle={2023 IEEE/RSJ International Conference on Intelligent Robots and Systems (IROS)},
  year={2023},
}

@article{ruder2017overview,
  title={An overview of multi-task learning in deep neural networks},
  author={Ruder, Sebastian},
  journal={arXiv:1706.05098},
  year={2017}
}

@article{sutton1988td,
  title={Learning to predict by the methods of temporal differences},
  author={Sutton, Richard S.},
  journal={Machine Learning},
  year={1988},
}
}

\end{document}